\documentclass[sigconf]{acmart}
\AtBeginDocument{%
  }
\definecolor{myred}{rgb}{0.7, 0, 0} 
\definecolor{myblue}{RGB}{216,27,96}
\definecolor{mygreen}{RGB}{30,136,229}
\definecolor{mygold}{RGB}{44, 130, 4}
\definecolor{mypurp}{RGB}{93,58,155}

\definecolor{mylime}{RGB}{225,195,106}
\newcommand{\gr}[1]{\textcolor{mygreen}{#1}}
\newcommand{\bl}[1]{\textcolor{myblue}{#1}}
\newcommand{\gl}[1]{\textcolor{mygold}{#1}}
\newcommand{\pur}[1]{\textcolor{mypurp}{#1}}

\theoremstyle{definition}
\newtheorem{remark}{Remark}
\usepackage{tikz}

\newcommand{\BM}{\begin{matrix}}
\newcommand{\EM}{\end{matrix}}

\newcommand{\PP}{\mathbb{P}}
\newcommand{\CC}{\mathbb{C}}
\newcommand{\RR}{\mathbb{R}}
\newcommand{\FF}{\mathbb{F}}
\newcommand{\QQ}{\mathbb{Q}}
\newcommand{\ZZ}{\mathbb{Z}}

\usepackage{arydshln}

\newcommand{\qq}[1]{\bl{\mathbf{q}_{#1}}}
\newcommand{\pp}[1]{\textcolor{mygold}{\mathbf{p}_{#1}}}
\newcommand{\cam}[1]{\gr{\mathbf{A}_{#1}}}

\def\acts{\curvearrowright}

\usetikzlibrary{patterns}
\usetikzlibrary{positioning}
\usepackage{tikz-3dplot}
\usepackage{cleveref}

\setcopyright{acmlicensed}
\copyrightyear{2025}
\acmYear{2025}
\acmDOI{XXXXXXX.XXXXXXX}
\acmConference[ISSAC '25]{ISSAC 2025}{July 28--August 1,
  2025}{Guanjuato, Mexico}
\acmISBN{978-1-4503-XXXX-X/2025/06}

\usepackage{listings}
\definecolor{codedarkgreen}{RGB}{51, 133, 4}
\definecolor{codemaroon}{RGB}{133, 5, 63}
\definecolor{codeteal}{RGB}{0, 145, 109}
\definecolor{codepurple}{RGB}{123, 35, 125}

\lstdefinelanguage{Macaulay2}{
basicstyle=\normalsize\ttfamily,
  alsoletter=",
  classoffset=1,
  keywords={p0,p1,p2,p3,p4,p5,p6,p7,p8,p9,p10,p11,p12,p13,p14,p15,p16,p17,p18,p19,gap,o12,input,random,fold,setRandomSeed,norm,apply,compress,toList,vars,entries,matrix,transpose,det,subsets,genericMatrix,needsPackage,presentation,generators,gens,precision,selectInSubring,i1,i2,i3,i4,i5,i6,i7,i8,i9,i10,i11,flatten,ideal,GaloisWidth,local,if,elif,then,not,else,return,fi,end,first,reshape,max},
  keywordstyle={\color{blue}},
classoffset=2,
breaklines=true,
morekeywords={"MonodromySolver","Engine",output,PermList,Read,StructureDescription,Blocks,Group,Maximum,Kernel,IsPrimitive,OrbitsDomain,IsCylic,Length,IsTrivial,IsNaturalSymmetricGroup,IsNaturalAlternatingGroup,Facgors,Order,List,Orbits,Image,ActionHomomorphism,Factors,IsCyclic,IsTransitive,Representative,ConjugacyClassesMaximalSubgroups,Minimum,CompositionSeries,IsSimple},
keywordstyle={\color{codemaroon}},
classoffset=3,
morekeywords={CC,QQ,RR,CacheTable,Matrix,Eliminate},
keywordstyle={\color{codedarkgreen}},
classoffset=4,
morekeywords={restart,id_,false,true,Weights,Limit,Lex,MonomialOrder,FileName,OnSets,function},
keywordstyle={\color{codeteal}},
classoffset=5,
morekeywords={list,for,in,from,to,of},
keywordstyle={\color{codepurple}},
xleftmargin=1em,
xrightmargin=1em,
columns=fullflexible,
keepspaces=true,
stepnumber=1,
numbers=none,
captionpos=b,
showspaces=false,
frame=none
}

\begin{document}

\title{Numerically Computing Galois Groups of Minimal Problems}

\author{Timothy Duff}
\email{tduff@missouri.edu}
\orcid{https://orcid.org/0000-0003-2065-6309}
\affiliation{%
  \institution{University of Missouri - Columbia}
  \city{Columbia}
  \state{Missouri}
  \country{USA}
}

\renewcommand{\shortauthors}{T.~Duff}

\begin{abstract}
I discuss a seemingly unlikely confluence of topics in algebra, numerical computation, and computer vision. The motivating problem is that of solving multiples instances of a parametric family of systems of algebraic (polynomial or rational function) equations. No doubt already of interest to ISSAC attendees, this problem arises in the context of robust model-fitting paradigms currently utilized by the computer vision community (namely "Random Sampling and Consensus", aka "RanSaC".) This talk will give an overview of work in the last 5+ years that aspires to measure the intrinsic difficulty of solving such parametric systems, and makes strides towards practical solutions.
\end{abstract}


\maketitle

\section{Introduction}

This article accompanies an invited tutorial presented at the ISSAC 50th anniversary conference in Guanajuato. 
It offers expository accounts of algebraic vision~\cite{kileel2022snapshot}, Galois groups of polynomial systems~\cite{sottile2021galois}, and numerical algebraic geometry~\cite{bates2023numerical}.
Needless to say, the 8-page format prevents treating any one of these topics in-depth. I encourage readers seeking more details to look at the surveys cited above. 
The main novelty is the synthesis of these three topics.

\section{Algebraic Vision}\label{sec:min-probs}

Every application of algebraic geometry I've ever worked on is motivated by some instance of the following ``universal problem": \emph{Let $\bl{\pi } : \gr{\mathcal{X}} \dashrightarrow \gl{\CC^m} $ be a rational map, $\gr{\mathcal{X}}$ a variety of unknown states, and $\bl{\pi} (\gr{\mathcal{X}}) \subset \gl{\CC^m}$  a space of idealized data.
Given a measurement $\gl{\widetilde{y}} \approx \gl{y}$ of ``true" data $\gl{y} = \bl{\pi } (\gr{x})$, recover $\gr{\widetilde{x}} \in \gr{\mathcal{X}}$ with $\gr{\widetilde{x}} \approx \gr{x}$.}

Computer vision is replete with such problems.
I work over the complex numbers for simplicity.
This is bearing in mind, of course, that in practice the input $\gl{\widetilde{y}}$ is real-valued, and that your solution $\gr{\widetilde{x}}$ also better be real if you want anyone to care about your work.

\begin{example}\label{ex:pnp}
The classic ``Perspective-$n$-Point" problem (P$n$P) can be formulated as follows: fix $\qq{1}, \ldots , \qq{n} \in \bl{\PP \left( \CC^{4\times 1}\right)},$ and define
\begin{align}\label{eq:p3p-map}
\bl{\pi} : \gr{\operatorname{SE}_3} &\dashrightarrow \gl{\left(\CC^2 \right)^n}\\
\gr{\left(\mathbf{R} \mid \mathbf{t}\right)}
&\mapsto 
\left(\Pi \left( \gr{\left(\mathbf{R} \mid \mathbf{t}\right)} \qq{1} \right), \, \ldots , \, \Pi  \left( \gr{\left(\mathbf{R} \mid \mathbf{t}\right)} \qq{n} \right)\right) \nonumber \\
&\text{where} \quad  \Pi (x,y,z) = (x/z, y/z). \nonumber 
\end{align}
The data $\qq{1}, \ldots , \qq{n}$ represent three-dimensional points in homogeneous coordinates, and $\gr{\operatorname{SE}_3}$ is the (complexified) special Euclidean group.
Here is the problem: given $\gl{\mathbf{y}_1}, \ldots ,\gl{\mathbf{y}_n} \in \gl{\CC^2}$ and the corresponding $\qq{1},\ldots , \qq{n},$ find $\gr{\left(\mathbf{R} \mid \mathbf{t}\right)}\in \gr{\operatorname{SE}_3}$ such that 
\begin{equation}\label{eq:pnp-inexact}
\bl{\pi} \left( \gr{\left(\mathbf{R} \mid \mathbf{t}\right)} \right) \approx (\gl{\mathbf{y}_1}, \ldots , \gl{\mathbf{y}_n}).
\end{equation}

Any discussion of how to define the symbol $\approx $ would lead us too far astray, so let's consider a scenario that is quite natural for the ISSAC community: the existence of an \emph{exact solution},
\begin{equation}\label{eq:exact-pnp}
\bl{\pi} \left( \gr{\left(\mathbf{R} \mid \mathbf{t}\right)} \right) = (\gl{\mathbf{y}_1}, \ldots , \gl{\mathbf{y}_n}).
\end{equation}
Counting dimensions, it should come as little surprise that an exact solution exists when $n=3$ \emph{for generic data} $(\qq{1} , \qq{2}, \qq{3}, \gl{\mathbf{y_1}}, \gl{\mathbf{y_2}}, \gl{\mathbf{y_3}})$. 
This is the hallmark of a \emph{minimal problem}.

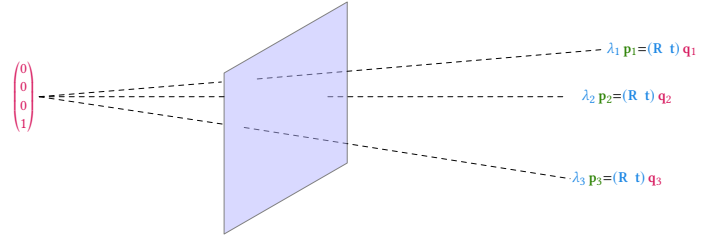
\begin{figure}
    \centering
    \resizebox{1.1\columnwidth}{!}{
        \begin{tikzpicture}[line join = round, line cap = round]
        \coordinate[label=left:$\bl{\begin{pmatrix} 0 \\ 0\\0\\1\end{pmatrix}}$] (O) at (-5.2,0,0);
        \coordinate (a1) at (-5 +3.64279, .3105, 0);
        \coordinate (a2) at (-5 + 4.3995, .375, 0);
        \coordinate[label=right:$\hspace{-.2em} \gr{\lambda_1}  \, \pp{1} \text{$=$} \gr{\left( \mathbf{R} \, \, \,  \mathbf{t} \right)} \, \qq{1}$] (a3) at (-5 + 11.732, 1, 0);
        \draw[dashed] (a2) -- (a3);
        \coordinate (b1) at (-5 + 3.7455, -.589746, 0);
        \coordinate (b2) at (-5 + 4.125, -.6495, 0);
        \coordinate[label=right:$\hspace{-.2em} \gr{\lambda_3}  \, \pp{3} \text{$=$} \gr{\left( \mathbf{R} \, \, \,  \mathbf{t} \right)} \, \qq{3}$] (b3) at (-5 + 11, -1.732, 0);
        \draw[dashed] (b2) -- (b3);
        \coordinate (c1) at (-1.24,0,0);
       \coordinate (c2) at (.89,0,0);
        \node[circle,label=right:$\hspace{-.2em} \gr{\lambda_2}  \, \pp{2} \text{$=$} \gr{\left( \mathbf{R} \, \, \,  \mathbf{t} \right)} \, \qq{2}$] (c3) at (6.05,0,0) {};
        \draw[dashed] (c2) -- (c3);
        \coordinate  (TL) at (-1.3,.5,0);
        \coordinate  (TR) at (1.298, 2, 0);
        \coordinate  (BL) at (-1.3, -2.9, 0);
        \coordinate  (BR) at (1.298, -1.4, 0);
        \foreach \i in {a1,b1,c1}
            \draw[dashed] (O)--(\i);
        \draw[-, fill=blue!30, opacity=.5] (TL)--(TR)--(BR)--(BL)--cycle;
    \end{tikzpicture}  
    }
    \caption{Illustration of Perspective-$3$-Point: $\pp{i} $ denotes the normalized homogeneous coordinates of a 2D point $\gl{\mathbf{y}_i} \in \gl{\RR^2}$, and $\gr{\lambda_i}$ denotes the projective depth of the 3D point $\qq{i}$ (in affine terms, the distance from $\qq{i}$ to the center of projection.)}\label{fig:pnp}
\end{figure}

P3P, all things considered, is a pretty simple problem, and also a pretty old one, dating back to Lagrange in the 1700s (see~\cite{sturm2011historical} for related history.)
Thus, it is remarkable that P3P continues to receive considerable attention in recent literature coming from both symbolic computation~\cite{DBLP:conf/issac/GaillardD24,DBLP:conf/issac/FaugereMRD08} and computer vision~\cite{DBLP:conf/eccv/PerssonN18,ding23}.

Minimal problems might seem like a theoretical curiosity---after all, wouldn't you really like to solve the inexact problem~\eqref{eq:pnp-inexact}, for any $n$?
However, the computer vision community has realized that minimal problems can be a surprisingly effective tool for outlier-robust estimation.
To motivate this, note that the inexact formulation of~\eqref{eq:pnp-inexact} is still rather naive, since it assumes the knowledge of \emph{exact point correspondences.}
In reality, these correspondences usually arise as byproducts of image processing or machine learning methods, and as a result are not merely noisy, but often flat out \emph{wrong.}

To cope with erroneous correspondences, we could try multiple trials in which we solve~\eqref{eq:pnp-inexact} for different sub-samples of the data; then, using the solutions obtained in each trial, we may attempt to discern which data are erroneous.
This is the essential idea behind a popular paradigm known as \emph{Random Sampling and Consensus}, aka \emph{RanSaC}, introduced by Fischler and Bolles in 1981~\cite{fischler1981random}.
Following the more recent work~\cite{DBLP:journals/corr/abs-2503-07829}, we analyze the complexity of RanSaC in its simplest of forms. 
Assume we have the following:
\begin{enumerate}
\item $N$, the number of trials,
\item a working solver for the P$k$P problem, where $k$ is the number of correspondences used in any of the $N$ trials,
\item a ``consensus" criterion by which unsampled correspondences are judged to be either consistent or inconsistent with a solution (``inliers" and ``outliers", respectively),
\item $s\in (0,1),$ the desired probability of obtaining an outlier-free sample of $k$ correspondences after $N$ trials.
\end{enumerate}
Let $p\in (0,1)$ denote the fraction of erroneous correspondences, so that the probability of drawing an all-inlier sample in one trial is 
\begin{equation}\label{eq:p-success-one-trial}
P = \displaystyle\frac{\binom{pn}{k}}{\binom{n}{k}}.
\end{equation}
From our specification above, we should have
\begin{equation}\label{eq:probability-equation}
(1-P)^N \le 1 - s,
\end{equation}
so the number of trials should satisfy
\begin{equation}\label{eq:number-of-trials}
N \ge \displaystyle\frac{\log (1-s)}{\log (1-P)}.
\end{equation}

\begin{figure}
    \centering
    \includegraphics[width=\linewidth]{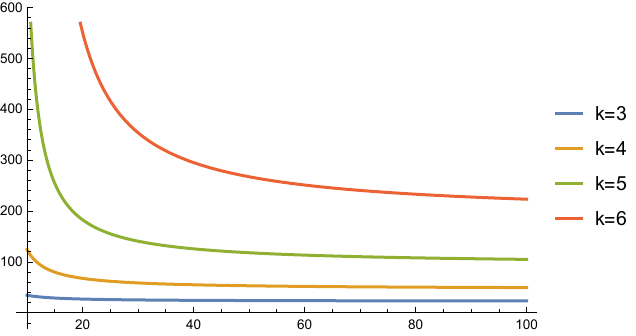}
    \caption{Based on~\eqref{eq:number-of-trials}, the number of RanSaC trials $N$ needed to find an outlier-free subsample of size $k$ with $95\%$ confidence from $n \in [10,100]$ total correspondences, with $50$\% outliers. }
    \label{fig:ransac-plot}
\end{figure}
\Cref{fig:ransac-plot} plots the right-hand side of~\eqref{eq:number-of-trials} for $s=.95,$ $p=.5,$ $n \in \{ 10, \ldots , 100 \}$ and $k\in \{ 3,4,5 ,6 \}.$
Clearly, fewer trials are needed as we decrease the size of subsamples $k.$
In the case of P3P, $k=3$ is the absolute minimum number of correspondences we'd need to solve the P$k$P problem up to finitely-many candidate solutions.
Typically, an instance of the P3P problem has more than one plausible solution.
Thus, each trial generates multiple candidates that must scored accordingly inside of the RanSaC loop in order to obtain an unambiguous estimate of the ``true" solution.
This solution is then refined after running RanSaC, using nonlinear least squares with the associated consensus set.
In fact, RanSaC exists in a multitude of different flavors, such as LO-RanSaC~\cite{DBLP:conf/dagm/ChumMK03}, which incorporates such refinement steps into each of the trials.
I couldn't possibly do justice to the vast landscape of different RanSaCs, especially in this short article.
Still, 
regardless of the RanSaC flavor that one uses, the point remains that \emph{the most sample-efficient solvers are minimal.}

Let's return to P3P---how might we actually solve it?
Referring to~\Cref{fig:pnp}, we'll put each 2D point $\gl{\mathbf{y}_i} \in \gl{\RR^2}$ into homogeneous coordinates, $\pp{i} \in \gl{\PP \left( \mathbb{R}^{3\times 1}\right)}$, \emph{normalized} so that $\pp{i}^T \pp{i} = 1.$
If $\qq{1}, \qq{2}, \qq{3}$ are expressed in the affine chart where their last coordinates equal 1, then we must have
\begin{equation}\label{eq:p3p-full}
\gr{\lambda_i} \pp{i} = \gr{\left(\mathbf{R} \mid \mathbf{t}\right)} \qq{i},
\quad i=1,2,3,
\end{equation}
for some unknown scalar \emph{depths} $\gr{\lambda_1}, \gr{\lambda_2},\gr{\lambda_3}.$ 
Using the law of cosines, one can obtain a system of three equations in the unknown depths,
\begin{align}
    \gr{\lambda_1}^2 +
\gr{\lambda_2}^2
-
2\left(\pp{1}^T \pp{2}
\right)
\gr{\lambda_1} \gr{\lambda_2} =
(\qq{1}-\qq{2})^T (\qq{1}-\qq{2}), \nonumber  \\
    \gr{\lambda_1}^2 +
\gr{\lambda_3}^2
-
2\left(\pp{1}^T \pp{3}
\right)
\gr{\lambda_1} \gr{\lambda_3} =
(\qq{1}-\qq{3})^T (\qq{1}-\qq{3}), \nonumber \\
    \gr{\lambda_2}^2 +
\gr{\lambda_3}^2
-
2\left(\pp{2}^T \pp{3}
\right)
\gr{\lambda_2} \gr{\lambda_3} =
(\qq{2}-\qq{3})^T (\qq{2}-\qq{3}) . \label{eq:grunert}
\end{align}
The system~\eqref{eq:grunert} generically has the ``expected" $2^3=8$ solutions.

Intersecting three quadrics in three unknowns is a special, well-studied problem~\cite{DBLP:conf/cvpr/KukelovaHF16}, but P3P is \emph{even more} special.
This is because solutions to~\eqref{eq:grunert} exist in symmetric pairs, $(\pm \gr{\lambda_1}, \pm \gr{\lambda_2}, \pm \gr{\lambda_3}).$
It's also quite reasonable to assume that the depths are nonzero.
Introducing new unknowns $(\gr{\rho_1}, \gr{\rho_2}) = (\gr{\lambda_1} / \gr{\lambda_3}, \gr{\lambda_2}/{\gr{\lambda_3}})$, we deduce from~\eqref{eq:grunert} that
\begin{align}
\bl{d_{13}} \left(
\gr{\rho_1}^2 + \gr{\rho_2}^2 + \gl{c_{12}} \gr{\rho_1} \gr{\rho_2} 
\right) &= \bl{d_{12}} \left(
1 + \gr{\rho_1}^2  + \gl{c_{13} \gr{\rho_1}} 
\right) \nonumber , \\
\bl{d_{23}} \left(
\gr{\rho_1}^2 + \gr{\rho_2}^2 + \gl{c_{12}} \gr{\rho_1} \gr{\rho_2} 
\right) &= \bl{d_{12}} \left(
1 + \gr{\rho_2}^2  + \gl{c_{23}} \gr{\rho_2} 
\right), \label{eq:p3p-conics}
\end{align}
where $\gl{c_{ij}} = \pp{i}^T \pp{j}$ and $\bl{d_{ij}} = (\qq{i}-\qq{j})^T (\qq{i}-\qq{j})$. 
In other words, P3P reduces to intersecting \emph{two plane conics!}

How is Galois theory relevant to this example?
Well, two generic plane conics intersect in four points, and if these conics are defined over $\QQ ,$ then the intersection points have coordinates which are algebraic numbers of degree $4.$
In principle, these coordinates could be represented exactly on a computer by their quartic minimal polynomials. 
On the other hand, Galois theory provides us with the following mantra: \emph{solving the quartic can be achieved by solving an auxiliary cubic and auxiliary 
quadratics.} 
For P3P, this mantra manifests as follows: consider the \emph{conic pencil}
\begin{equation}\label{eq:conic-pencil}
\begin{pmatrix}
\gr{\rho_1} &
\gr{\rho_2} & 
1
\end{pmatrix}
\left( 
\pur{t} C_1 + (1-\pur{t}) C_2 
\right)
\begin{pmatrix}
\gr{\rho_1} \\
\gr{\rho_2} \\ 
1
\end{pmatrix}
=0,
\end{equation}
where $C_1$ and $C_2$ are $3\times 3$ symmetric matrices representing the two conics~\eqref{eq:p3p-conics}. 
By genericity, we may assume $C_2$ is invertible;
if the two conics intersect at a point $(\gr{\rho_1}, \gr{\rho_2})$ in the pencil, then the corresponding value of $\pur{t}$ must solve the $3\times 3$ eigenvalue problem 
\begin{equation}\label{eq:pencil-eigenvalue}
\det \left( 
I - \pur{t} \left( \left( C_2 - C_{1}\right)^{-1} C_2 \right)
\right) = 0.
\end{equation}
Equation~\eqref{eq:pencil-eigenvalue} is our auxiliary cubic.
For each of its roots $\pur{t}$, we may form the rank-2 symmetric matrix
\begin{equation}\label{eq:rank-2-matrix}
\pur{t} C_1 + (1-\pur{t}) C_2,
\end{equation}
representing a degenerate conic. Generically, this is the union of two lines which can be found using the formulas of~\cite[Ch.~12]{richter-gebert}; these involve taking the adjugate of~\eqref{eq:rank-2-matrix} and working in a suitable quadratic extension of the field $\QQ (\pur{t})$.
Intersecting these two lines with one of the original conics $C_i$ recovers the four intersection points $(\gr{\rho_1}, \gr{\rho_2}).$
At last, one more quadratic extension recovers the unknown depths $(\gr{\lambda_1}, \gr{\lambda_2},\gr{\lambda_3}),$ and from there it is only a matter of simple algebra to recover the unknowns $\gr{\mathbf{R}}$ and $\gr{\mathbf{t}}$. 
This type of ``cubic solution" to P3P has long been known---see~\cite{DBLP:journals/ijcv/HaralickLON94} for a review of early work, and~\cite{DBLP:conf/eccv/PerssonN18,ding23} for more recent developments.
\end{example}

The name \emph{algebraic vision}~\cite{kileel2022snapshot} has recently been coined to describe the interface between algebraic geometry (often of a computational flavor) and problems like P3P which arise in computer vision.
This algebraic perspective has produced novel insights and improved solutions to many classical instances of the previously-described ``universal problem": for example, \emph{camera resectioning / absolute pose}~\cite{connelly2024algebra,hruby2024efficient,DBLP:conf/eccv/KukelovaASSP20}, \emph{3D point triangulation}~\cite{DBLP:conf/cvpr/KukelovaL19,maxrodwang,duff2024metric}, \emph{3D reconstruction / relative camera pose}~\cite{kiehn2025plmp,DBLP:conf/cvpr/HrubyKDOPPL23,DBLP:conf/iccv/ArrigoniPF23,DBLP:journals/corr/abs-2310-02719,DBLP:journals/siaga/Kileel17,plmp-duff,hruby2023learning,DBLP:conf/cvpr/HrubyKDOPPL23}, and \emph{autocalibration}~\cite{DBLP:conf/cvpr/CinDMP24,DBLP:conf/cvpr/KocurKK24,DBLP:conf/cvpr/KukelovaP07}.
From a theoretical angle, it is interesting to answer fundamental algebro-geometric questions about these problems and how they relate; see~\cite{agarwal2022atlas} for one recent overview.
From a more practical angle, 
algebraic vision derives much of its mandate from the study of minimal problems, which may lead to more efficient minimal solvers.

The P$n$P problem is nothing more than the resectioning problem for a calibrated perspective camera, given 3D-2D point matches.
This is generalized in the next example.
\begin{example}\label{ex:resect-lines}
Given $\gl{p}$ 3D-2D point matches and $\pur{l}$ 3D-2D \emph{line} matches, the map defining the calibrated resectioning problem is
\begin{align}
\gl{\pi '} : \gr{\operatorname{SE}_3} &\dashrightarrow 
\gl{( \gl{\PP^2} )^{p}} \times \pur{\operatorname{Gr} (\PP^1, \PP^2)^{l}} \label{eq:gen-resect-map} \\
\gr{\left( \mathbf{R} \, \, \,  \mathbf{t} \right)} 
&\mapsto \left( 
\gr{\left( \mathbf{R} \, \, \,  \mathbf{t} \right)} \qq{1},  
\ldots , 
\gr{\left( \mathbf{R} \, \, \,  \mathbf{t} \right)} \qq{p},
\gr{\wedge^2 \left( \mathbf{R} \, \, \,  \mathbf{t} \right)} \pur{\mathbf{\ell}_1}
,
\ldots 
,
\gr{\wedge^2 \left( \mathbf{R} \, \, \,  \mathbf{t} \right)} \pur{\mathbf{\ell}_l}
\right). \nonumber 
\end{align}
where $\gr{\wedge}$ denotes the exterior square of a linear map, and we have fixed $\qq{1}, \ldots , \qq{p} \in \bl{\PP^3}$ (as in P$n$P) and lines $ \, 
\mathbf{\pur{\ell_1}} , 
\ldots \mathbf{\pur{\ell_l}} \in \pur{\operatorname{Gr} (\PP^1 , \PP^3)}.$
This generalized resectioning problem is minimal when $\gl{p} + \pur{l} = 3,$ and $(\gl{p}, \pur{l}) = (\gl{3}, \pur{0})$ is just P3P in projective coordinates.
In the computer vision literature, the minimal cases involving lines have been studied in a number of previous works.

Solvers for the ``pure lines" case $(\gl{0}, \pur{3})$ based on the intersection of three quadrics date back to the 1980s~\cite{DBLP:journals/pami/DhomeRLR89}.
In comparison with P3P, it is natural to ask: can we also reduce to the problem of intersecting two conics?
This is indeed possible when the lines are in certain special positions (eg.~three lines meeting at a point~\cite{DBLP:journals/pami/XuZCK17}.)
However, analyzing the Galois group associated to the problem, which turns out to be the full-symmetric group $S_8$~\cite{galois-siaga,hruby2024efficient}, shows that \emph{such a reduction is generally impossible}.

The ``mixed cases" $(\gl{2}, \pur{1})$ and $(\gl{1}, \pur{2})$ have also been studied. 
For example,~\cite{DBLP:conf/icra/RamalingamBS11} proposes a pipeline that uses such solvers in the context of urban localization tasks.
These solvers were already quite efficient, but were later significantly improved in~\cite{hruby2024efficient}.
These improvements exploited the fact that both of these problems \emph{decompose} into subproblems, as suggested by Galois group computation.
\end{example}

We now move from resectioning onto harder problems.

\begin{example}\label{ex:recon}
\begin{figure}
    \centering
 \def\t{0.225}
 \def\s{1.8}
 \def\u{-0.0875}
 \def\uu{-0.2755}
 \def\uuu{-0.655}
 \def\uuuu{0.15}
 \def\v{-0.1305}
 \begin{tikzpicture}[scale = 0.8]
 \coordinate (c) at (0,0,0);
 \coordinate (p) at (-1,3/2,0);
 \coordinate (x0) at (\t,-\t*3/2,0);
 \coordinate (x) at (1.02,-1.53,0);
 \coordinate (x1) at (\s,-\s*3/2,0);
 \coordinate (e1) at (1,0,0);
 \draw[fill=mygreen!10] (1,-2,-2) -- (1,-2,2) -- (1,2,2) -- (1,2,-2) node[right] {} -- cycle;
 \draw[thick, dashed, -] (p) -- (x0) ;
 \draw[thick, dashed, ->] (x) -- (x1) ;
 \draw[fill=myblue] (p) circle (0.05) node[above] {$\qq{i}$};
 \draw[fill=mygold] (x) circle (0.05) node[above] {$\pp{2 i}$};
 \draw[fill=mygreen] (0,0,0) circle (0.05) node[left] {$\gr{\begin{pmatrix}
 -\mathbf{R_2}^T \gr{\mathbf{t_2}}  \\
 1
 \end{pmatrix}}$};
 \coordinate (c2) at (-5,0,0);
 \coordinate (e2) at (-5.95,0,0);
  \draw[fill=mygreen!10] (-6,0,-5/2) -- (-6,-5/2,0)  -- (-6,0,5/2) --  (-6,5/2,0) node[right] {} -- cycle;
 \coordinate (x2) at (4*\u-5, 3/2*\u, 0);
 \draw[thick, dashed, -] (p) -- (x2) ;
 \coordinate (x3) at (4*\uu-5, 3/2*\uu, 0);
 \coordinate (x4) at (4*\uuu-5, 3/2*\uuu, 0);
  \coordinate (h) at (-5.59,0, 0);
 \draw[thick, dashed, ->] (x3) -- (x4) ;
 \draw[fill=mygold] (x3) circle (0.05);
 \node (cap) at ((4*\uu-5, -0.8, 0) {$\pp{1 i}$};
 \coordinate (x5) at (4*\uuuu-5, 3/2*\uuuu, 0);
 \draw[fill=mygreen] (x5) circle (0.05) node[below right] {
$\gr{\begin{pmatrix}
 \mathbf{0} \\
 1
 \end{pmatrix}}$};
 \end{tikzpicture}    
    \caption{Geometry of the calibrated stereo pair~\eqref{eq:cams-fixed}.}
    \label{fig:stereo}
\end{figure}
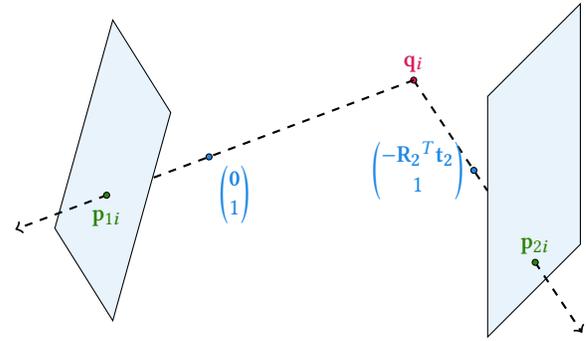
The \emph{reconstruction problem} asks us to recover $\bl{n}$ 3D points and $\gr{m}$ cameras from  projections.
Unlike resectioning, $\gr{m} \ge \gr{2}$ cameras are needed in order to solve this problem.
This is for the same reason humans need two eyes to perceive depth; see~\Cref{fig:stereo} for an illustration of the $\gr{m}=\gr{2}$ ``stereo vision" problem.
To simplify the exposition, we assume each camera is \emph{calibrated}---that is, an element $\cam{} \in \gr{\operatorname{SE}_3}$.
Thus, inverting the map
\begin{align}
\label{eq:recon-map}
\Psi_{\gr{m}, \bl{n}} : 
\gr{\operatorname{SE}_3^m} \times  \bl{(\PP^3)^n} &\dashrightarrow \gl{\left(\PP^2 \right)^{mn}}\\
(\cam{1}, \ldots , \cam{m}, \qq{1}, \ldots , \qq{n})  &\mapsto (\cam{1} \qq{1}, \ldots , \cam{m} \qq{n} ).
\end{align}
seems to be a natural formulation for the reconstruction problem.
However, for any $\gr{m}$ and $\bl{n}$, the map~\eqref{eq:recon-map} has positive-dimensional fibers, due to the action of the \emph{similarity group}
\begin{equation}
\begin{split}
\pur{\mathcal{S}_3} = \left\{ \pur{\mathbf{H}} \in \pur{\operatorname{PGL}_4} \mid 
\pur{\mathbf{H}} \sim 
\pur{\begin{pmatrix}
\mathbf{R} & \mathbf{t} \\
0 & s
\end{pmatrix}
},
\quad 
\pur{\left( \mathbf{R} \, \vert \, \mathbf{t}\right)} \in \pur{\operatorname{SE}_3},
\quad 
\pur{s} \in \pur{\CC^*}
\right\}\label{eq:sim-grp},\\
\pur{\mathbf{H}} \cdot (\cam{1}, \ldots , \cam{m}, \qq{1}, \ldots , \qq{n})
=
(\cam{1}\pur{\mathbf{H}}^{-1}, \ldots , \cam{m} \pur{\mathbf{H}}^{-1}, \pur{\mathbf{H}}\qq{1}, \ldots , \pur{\mathbf{H}}\qq{n}). \nonumber 
\end{split}
\end{equation}
(Here, and later, $\sim$ denotes equality up to scale.)

The ambiguity of solutions created by this group action is a natural one: reconstruction from calibrated cameras is only possible up to a choice of world coordinates and global scaling.
Thus, we may reformulate reconstruction as inverting the rational map 
\begin{align}
\label{eq:recon-map-quo}
\widetilde{\Psi_{\gr{m}, \bl{n}}} : 
\left( \gr{\operatorname{SE}_3^m} \times  \bl{(\PP^3)^n}\right) / \pur{\mathcal{S}_3} &\dashrightarrow \gl{\left(\PP^2 \right)^{mn}}\\
(\cam{1}, \ldots , \cam{m}, \qq{1}, \ldots , \qq{n})  &\mapsto (\cam{1} \qq{1}, \ldots , \cam{m} \qq{n} ),
\end{align}
where the quotient may be understood as the ``rational quotient" of Rosenlicht~\cite{MR82183}.
A typical trick used to model this quotient is to fix the first two cameras as follows:
\begin{equation}\label{eq:cams-fixed}
\cam{1} \sim \gr{\left( \mathbf{I} \, \vert \, \mathbf{0} \right)} ,
\quad 
\cam{2} \sim 
\gr{\left( \mathbf{R_2} \, \vert \, \mathbf{t}_2 \right)},
\quad 
\gr{\mathbf{t}_2} = \gr{\begin{pmatrix}
t_1\\
t_2\\
1
\end{pmatrix}}.
\end{equation}
A minimal case for inverting~\eqref{eq:recon-map-quo} occurs when $(\gr{m}, \gl{n}) = (\gr{2}, \bl{5})$: this is expected, since both domain and codomain have dimension 
\[
\gr{2\cdot 6} + \bl{5 \cdot 3} - \pur{7}
= 
\gl{2\cdot 5 \cdot 2} = 20.
\]
In fact, a similar count shows that this is the \emph{only} minimal case for inverting $\widetilde{\Psi_{\gr{m}, \bl{n}}}.$ Reconstructing configurations of points and lines, however, has a much richer story (see eg.~\cite{plmp-duff,DBLP:journals/ijcv/DuffKLP24,DBLP:journals/pami/FabbriDFRPTWHGKLP23}).)

Let us explain the usual equations for reconstructing $\bl{5}$ points in $\gr{2}$ views, and how Galois theory enters the picture. 
Suppose 
\[
\Psi_{\gr{2}, \bl{5}} (\cam{1}, \cam{2}, \qq{1}, \ldots , \qq{5} ) = 
(\pp{11}, \ldots , \pp{25}),
\quad 
(\cam{1}, \cam{2}) \, \text{ as in~\eqref{eq:cams-fixed}.}
\]
For the stereo pair~\eqref{eq:cams-fixed}, we then have for $i=1,\ldots ,5$ that
\begin{equation}
\pp{1 i} \sim \cam{1} \qq{i}
\, \, 
\Rightarrow 
\, \, 
\pp{2 i}
\sim 
\cam{2} \qq{i} \sim 
\gr{\mathbf{R}} \pp{1 i} +  \gr{\mathbf{t}}
\, \, 
\Rightarrow 
\, \, 
\pp{2 i}^T \gr{\left( 
[\mathbf{t}]_\times 
\mathbf{R}
\right)}
\pp{1 i} = 0,
\label{eq:essential-matrix}
\end{equation}
where $\gr{[\mathbf{t}]_\times}$ is the $3\times 3$ skew symmetric matrix representing the cross-product as a linear map $\gl{\CC^3} \ni \pp{} \mapsto \gr{\mathbf{t}} \times \pp{} \in \gl{\CC^3}.$

The matrix $\gr{[\mathbf{t}]_\times 
\mathbf{R}}$ is known in computer vision as the \emph{essential matrix} associated to the stereo pair~\eqref{eq:cams-fixed}.
The \emph{Nist\'{e}r-Stew\'{e}nius five-point algorithm}~\cite{stewenius2006recent} reconstructs five points in three views by using the equations~\eqref{eq:essential-matrix} to derive a degree-$10$ polynomial, whose roots generically determine $10$ distinct complex-valued essential matrices.
Each essential matrix, in turn, gives rise to two possibilities for the stereo pair~\eqref{eq:cams-fixed}.
Finally, 3D points $\qq{i}$ are determined as the intersection of two lines as in~\Cref{fig:stereo}.
Thus, just like in~\Cref{ex:pnp,ex:resect-lines}, the five-point algorithm proceeds by decomposing the reconstruction problem into simpler algebraic subproblems.
This observation is plainly seen from the Galois group's action on the 20 solutions.
\end{example}

So far, we have exclusively treated calibrated pinhole cameras.
Algebraic vision, however, has a lot to offer to other camera models involving polynomial, rational, and even algebraic functions.
We conclude this section with one such example.

\begin{example}\label{ex:radial}
One reason to look beyond the pinhole camera model is the phenomenon of \emph{radial distortion,} in which the magnification of an image increases or decreases away from a fixed point (the \emph{center of distortion.})
A large number of models have been proposed for modeling radial distortion: for example, under the so-called \emph{division model}~\cite{DBLP:conf/cvpr/Fitzgibbon01}, we have
\begin{equation}\label{eq:radial-distortion}
\cam{} \qq{} \sim \pp{} \sim 
\pur{\begin{pmatrix}
y_1 \\
y_2 \\
1 + \gr{\rho} (y_1^2+ y_2^2)
\end{pmatrix}},
\end{equation}
where now $\pur{(y_1, y_2)} \in \pur{\CC^2}$ represents the measured 2D point, and $\gr{\rho}$ is a new, unknown \emph{radial distortion parameter.}
Note that, under this model, it is not straightforward to write the image point $\pur{(y_1, y_2)}$ as a function of the 3D point $\qq{}$ and the camera parameters $\cam{}, \gr{\rho}.$
However, if our true goal is to recover $\cam{}$ and $\qq{},$ then one can make use of the \emph{radial camera model}~\cite{DBLP:conf/cvpr/HrubyKDOPPL23}.
Here, we simply focus on the first two rows of~\eqref{eq:radial-distortion}, which do not involve $\gr{\rho }$: we have
\[
\gr{\widetilde{\cam{}}} \qq{} \sim \pur{\begin{pmatrix}
y_1 \\
y_2 
\end{pmatrix}},
\]
where $\gr{\widetilde{\cam{}}}  : \bl{\PP^3} \dashrightarrow \pur{\PP^1}$ is the \emph{calibrated radial camera} associated to the pinhole camera $\cam{}.$
Thus, calibrated radial cameras have the form
    \begin{equation}\label{eq:calib-radial-1}
\gr{\widetilde{\cam{i}}} = \gr{
\begin{pmatrix}
\mathbf{r}_{i1}^T & t_{i,1} \\
\mathbf{r}_{i2}^T & t_{i,2}
\end{pmatrix}
}
,
\quad 
\gr{\mathbf{r}_{i1}^T
\mathbf{r}_{i1}
} = \gr{\mathbf{r}_{i2}^T \mathbf{r}_{i2} } = 1,
\, \, 
\gr{\mathbf{r}_{i1}^T}\,\gr{\mathbf{r}_{i2}} = 0.
\end{equation}
Similarly to~\eqref{eq:recon-map-quo}, we may try to invert a reconstruction map
with codomain $\pur{(\PP^{1})^{mn}}$, and
whose domain consists $\pur{\mathcal{S}_3}$-orbits of pairs consisting $\gr{m}$-tuple of calibrated radial cameras and an $\bl{n}$-tuple of 3D points to their projections.
Analogously to~\eqref{eq:cams-fixed}, we we may assume
\begin{equation}\label{eq:calib-radial-2}
\gr{\widetilde{\cam{1}}} \sim  
 \gr{
\begin{pmatrix}
1 & 0 & 0 & 0\\
0 & 1 & 0 & 0
\end{pmatrix}
},
\quad 
\gr{\widetilde{\cam{2}}} \sim  
 \gr{
\begin{pmatrix}
\mathbf{r}_{21}^T & 0 \\
\mathbf{r}_{22}^T & 1
\end{pmatrix}
}.
\end{equation}
In this case, we haven't entirely resolved the $\pur{\mathcal{S}_3}$-ambiguity; the variety of all calibrated radial camera pairs $(\gr{\widetilde{A_1}} , \gr{\widetilde{A_2}})$ having the form~\eqref{eq:calib-radial-2}
is stabilized by the elements
$\pur{\mathbf{H}}=\pur{
\operatorname{diag} (1,1, \pm 1 , \pm 1)
}\in \pur{\mathcal{S}_3}$.

The first minimal case for radial camera reconstruction occurs when $(\gr{m}, \bl{n} ) =(\gr{4}, \bl{13})$: the problem is to solve
\begin{equation}\label{eq:radial-correspondence-constraints}
\gr{\widetilde{\cam{i}}} \bl{\mathbf{q}_j} \sim \pur{\mathbf{y}_{ij}} 
\quad 
\forall \, \gr{1\le i \le 4}, \, \bl{1\le j \le 13},
\end{equation}
where the data $\pur{\mathbf{y}_{ij}}\in \pur{\PP^1}$ are given. 
Taken together, equations~\eqref{eq:calib-radial-1},~\eqref{eq:calib-radial-2} and~\eqref{eq:radial-correspondence-constraints} are naturally formulated as a system of polynomials.
A numerical monodromy computation in~\cite{DBLP:conf/cvpr/HrubyKDOPPL23} determined this system to have 3584 solutions.
This might seem like terrible news.
However, as a byproduct of this same computation, one obtains a permutation group which serves as a good proxy for the true Galois group. 
The structure of this permutation group indicates that the solutions are highly structured: indeed, up to $\pur{\mathcal{S}_3}$-equivalence, there are merely $3584/16=224$ solutions.
What's more, the reconstruction problem decomposes further into sub-problems like the five-point problem of~\Cref{ex:recon}.
The natural analogue of the essential matrix for this problem is the \emph{radial quadrifocal tensor}: as it turns out, the $896$ $\pur{\mathcal{S}_3}$-orbits map onto $56=224/4$ $\pur{\operatorname{PGL}_4}$-orbits of \emph{uncalibrated} reconstructions, which in turn map onto $28$ radial quadrifocal tensors.
By carefully tracing through the associated subproblems, one may recover all 3584 solutions via a numerical homotopy continuation method that tracks just $28$ paths.
Needless to say, this is a significant improvement! We refer to~\cite{DBLP:conf/cvpr/HrubyKDOPPL23} for many more details.
\end{example}

Now that we've seen examples of minimal problems, let's give a general overview of their Galois groups and how to compute them.

\section{Galois groups of polynomial systems}
The minimal problems of the previous section all adhere to a common setup:
let $\gr{X} \subset \gl{\CC^m} \times \gr{\CC^n}$ be a variety whose points are \emph{\gl{problem}-\gr{solution} pairs}, for which we make the following assumptions:
\begin{enumerate}
    \item The projection $\bl{\pi } : \gr{X } \to \gl{\CC^m}$ onto the space of problems is dominant and generically $d$-to-$1$ for some $d\ge 1$, and
    \item The variety $\gr{X}$ is \emph{irreducible}.
\end{enumerate}
Solving a generic instance $\gl{z}\in \gl{\CC^m}$ of a minimal problem for us then simply means enumerating all $d$ solutions in the fiber $\bl{\pi}^{-1} (\gl{z}).$
We note that assumption 2, although seemingly restrictive, is satisfied by nearly all minimal problems encountered in practice.

It is often more convenient to describe the map $\bl{\pi}$ abstractly, with the understanding that the local description above is equivalent.
To be more precise, suppose we are given given an ``abstract" \emph{branched cover} $\bl{\pi '} : \gr{\mathcal{X}} \dashrightarrow \gl{\mathcal{Z}}$: that is, a dominant map  between irreducible complex algebraic varieties of dimension $\gl{m}.$ 
For minimal problems, one can produce explicit birational isomorphisms $\gr{t} : \gr{X} \dashrightarrow \gr{\mathcal{X}}$ and $\gl{b} : \gl{\CC^m} \dashrightarrow \gl{\mathcal{Z}}$ with $\bl{\pi '} \circ \gr{t} $ equal to $\gl{b} \circ \bl{\pi  }$ on a dense common domain of definition $\gr{U} \subset \gr{X}.$
In general, I would like to advocate the following philosophy: \emph{minimal problems are branched covers, and one should try to characterize their intrinsic complexity using birational invariants.}
For example, the ``abstract" P3P map $\gl{\pi '}$ of~\eqref{eq:gen-resect-map} with $(\bl{p}, \pur{l}) =(\gr{3}, \pur{0})$ is birationally-equivalent to a ``problem-solution pair" map $\gr{X} \to \bl{\CC^9} \times \gl{\CC^6}$ associated to the system~\eqref{eq:grunert}.
Similarly, the ``abstract" quotient map~\eqref{eq:recon-map-quo} has many equivalent models.

The overarching invariant, as you would probably expect, is the \emph{Galois group of the minimal problem}. 
Letting $\gr{K}$ denote the normal closure of the rational function field $\gr{\CC (X)}$ and $\gl{F}$ the function field of $\gl{\CC^m},$ this is simply the group $\gr{G}=\operatorname{Gal} (\gr{K} / \gl{F}).$

The Galois group encodes several more invariants of a minimal problem, via its guises as both an abstract group and as a permutation group acting on the solution set $\bl{\pi}^{-1} (\gl{z})$: mainly,
\begin{enumerate}
    \item[1.] The \emph{degree} / generic number of complex solutions $d$: when we compute $\gr{G}$ numerically, we typically obtain explicit permutations of the solution set. Abstractly, $d$ is the index of any \emph{point stabilizer} in $\gr{G}$ (which stabilizes a single solution.)
    \item[2.] The \emph{deck transformation group}, which may loosely be understood as the minimal problem's ``symmetry group"~\cite{DBLP:conf/issac/DuffKPR23}.
    Abstractly, this is the centralizer of $\gr{G}$ in $S_d.$
    For example, the P3P system~\eqref{eq:grunert}, the deck transformation group is $\gr{\ZZ / 2\ZZ},$ corresponding to the sign-symmetry $\gr{\lambda_i} \mapsto -\gr{\lambda_i}.$
   
    \item[3.] \emph{Decomposability}: does $\gl{\pi}$ factor as a composition of lower-degree maps?
    Using~\cite[Proposition 1]{brysiewicz}, this is equivalent to asking whether or not $\gr{G}$ acts \emph{imprimitively}, ie.~ whether it preserves a nontrivial partition of the solution set---or abstractly, whether point stabilizers are maximal subgroups.
    \item[4.] An invariant which I have recently proposed calling the \emph{Galois width}~\cite{duff2025galois}: this is the quantity 
\begin{equation}\label{eq:gal-width}
\gl{\operatorname{gw}} (\gr{G} ) = 
\gl{\displaystyle\min_{\substack{\gr{\text{unrefinable subgroup chains}}\\\gr{\text{id}}=\gr{H_m} \le \ldots \le \gr{H_0} = \gr{G}}}}
\left( 
\gl{\max_{0\le i < m}} \, \, [\gr{H_{i}} : \gr{H_{i+1}}]
\right).
\end{equation}
\end{enumerate}
The definition of the Galois width is motivated by a  simple, exact computation model: to compute an algebraic number $\gr{\alpha}$ in $\gl{m}$ steps,
\begin{enumerate}
    \item Initialize $\gr{\FF_0} \gets \gr{\QQ}$   
    \item For $\gl{i=1,\ldots , m},$ either  
    \begin{itemize}
        \item[(i)] do arithmetic in $\gr{\FF_i} \gets \gr{\FF_{i-1}},$ \gl{OR}  
        \item[(ii)] compute a root of a polynomial: formally, extend the working field $\gr{\FF_i} \gets \gr{\FF_{i-1}(\beta ) }$ 
    \end{itemize}
    \item Output: $\gr{\alpha } \in \gr{\FF_m}$ 
\end{enumerate}
It is natural to assign a cost of $[\gr{\FF_i} : \gr{\FF_{i-1}}]$ to each step of the algorithm.
If $\gr{G}$ is the Galois group of the minimal polynomial of $\gr{\alpha},$ then $\gl{\operatorname{gw}} (\gr{G} )$ measures the ``minimax" cost which minimizes the cost of the most costly step in any such algorithm computing $\gr{\alpha}$.~\cite{duff2025galois}

Computing Galois width is typically straightforward when $\gr{G}$ is known, due to the following results.

\begin{theorem}[Properties of Galois width (D '25)]
For any finite group $\gr{G}$, the following properties hold:
\begin{enumerate}
    \item $\gl{\operatorname{gw}} (\gr{H} ) \le \gl{\operatorname{gw}} (\gr{G} )$ for any subgroup $\gr{H} \le \gr{G}.$
    \item $\gl{\operatorname{gw}} (\gr{G} ) = \gl{\max \left( \gl{\operatorname{gw}} (\gr{N} ) , \gl{\operatorname{gw}} (\gr{G/N} ) \right)}$ whenever $\gr{N} \unlhd \gr{G}.$ 
    \item For any \emph{composition series}
$\gr{id} = \gr{N_m} \unlhd \gr{N_{m-1}} \unlhd \cdots \unlhd \gr{N_0} = \gr{G},$
    \[\gl{\operatorname{gw}} (\gr{G} ) = \gl{\max_{0\le i < m}} \gl{\operatorname{gw}}\left( \gr{N_i} / \gr{N_{i+1}} \right). \] 
     \item If $\gr{G}$ is simple, then
     \[
 \gl{\operatorname{gw}} (\gr{G} ) = \displaystyle\min_{\gr{H < G}} \, \, [\gr{G}: \gr{H}].
     \]  
      \item For any prime $\gr{p},$ we have $\gl{\operatorname{gw}} (\gr{\ZZ / p \ZZ} ) = \gr{p}.$ 
      \item For any $\gr{n}\ge 1,$ we have 
      $\gl{\operatorname{gw}} (\gr{S_n} ) = \gl{\operatorname{gw}} (\gr{A_n} ) = \begin{cases}
          \gr{3} \quad \text{if } \, \gr{n} = \gr{4}, \\
          \gr{n} \quad \text{else.}
      \end{cases}
      .$
\end{enumerate}
\end{theorem}    

\begin{figure}
    \centering
    \resizebox{\columnwidth}{!}{
 \begin{tikzpicture}[node distance=1cm]
\newcommand{\mydistance}{.6cm}
\node(S4)                     {$\gr{G}$};
\node(D8)       [below right=1cm and  .5cm of S4] {$(S_2 \times S_2 \times S_2) \rtimes D_8$};
\node(S3)      [below left=1cm and 1cm of S4]  {$(S_2 \times S_2 \times S_2) \rtimes S_3$};
\node(V4)      [below = 1cm of D8]  {$(S_2 \times S_2 \times S_2) \rtimes V_4$};
\node(C2)      [below = 1cm of S3]  {$(S_2 \times S_2 \times S_2) \rtimes S_3$};
\node (S22) [below = 1cm of V4] {$(S_2 \times S_2 \times S_2 ) \rtimes S_2$};
\node(V41)      [below = 1cm of S22]  {$S_2 \times S_2 \times S_2$};
\node(V42)      [below = 1cm of V41]  {$S_2 \times S_2 $};
\node(V43)      [below = 1cm of V42]  {$S_2$};
\node(id)      [below = 1cm of V43]  {$\operatorname{id}$};
\draw (S4) -- (D8) node [midway, fill=white] {3};
\draw (S4) -- (S3) node [midway, fill=white] {4};
\draw (D8) --(V4) node [midway, fill=white] {2};
\draw (V4) --(S22) node [midway, fill=white] {2};
\draw (S22) --(V41) node [midway, fill=white] {2};
\draw (V41) --(V42) node [midway, fill=white] {2};
\draw (S3) --(C2) node [midway, fill=white] {3};
\draw (V42) -- (V43) node [midway, fill=white] {2};
\draw (V43) -- (id) node [midway, fill=white] {2};
\draw (C2) -- (id) node [midway, fill=white] {16};
\end{tikzpicture}
}
    \caption{Two subgroup chains in the  Galois group of P3P.}
    \label{fig:gal-p3p}
\end{figure}
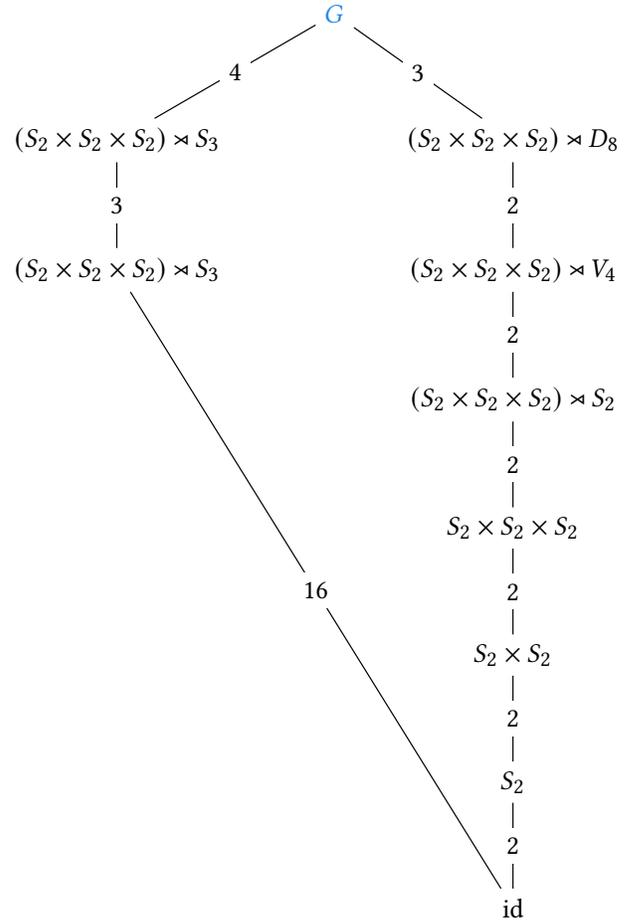

Galois width provides one (by no means the only) measure for the intrinsic difficulty of solving minimal problems. 
Computing it has already led in some cases to more efficient minimal solvers than previously known~\cite{hruby2023learning,hruby2024efficient}: it seems likely that this trend will continue.
Let's review the previous section's examples.

\begin{example}\label{ex:gwid-minimal}
The P3P problem of~\eqref{ex:pnp} has Galois width $\gr{3}.$ 
This observation relates to~\cite[\S 3]{galois-siaga}, where it was noted that the Galois group of this problem is the group of all even permutations of eight letters which preserve a fixed partition into four blocks of size two, $$\gr{G}=(S_2 \wr S_4)\cap A_8= (S_2\times S_2\times S_2) \rtimes S_4.$$
The chain of subgroups which attains the minimum in~\eqref{eq:gal-width} is shown on the right of~\Cref{fig:gal-p3p}.
The (refinable) chain on the left illustrates that chains of subgroups above the point stabilizers $$(S_2\times S_2) \rtimes S_3\le \gr{G}$$ are \emph{not} generally sufficient to determine $\gl{\operatorname{gw}}(\gr{G}).$  
In other words, the Galois width measures more than decomposability.

Moving along to the absolute pose problems of~\Cref{ex:resect-lines}: P3P1L has degree $4$ but Galois width $\gr{2},$ P1P2L has degree $8$ but Galois width $\gr{3}$, and P3L has both degree and Galois width $\gr{8}.$

The five-point relative pose problem of~\Cref{ex:recon} has degree $\gr{20}$ and Galois width $\gr{10},$ which shows that the Nist\'{e}r-Stew\'{e}nius method is algebraically optimal. 
This agrees with the results of~\cite{hartley}, which pioneered the computation of minimal problems' Galois groups.
In fact, the Galois groups used in that work were \emph{arithmetic} Galois groups, as opposed to the geometric Galois groups considered in this survey. 
These were computed by traditional methods of symbolic computation.
The relationship between these results is discussed in~\cite[\S 3]{duff2025galois}---suffice it to say that, even if one is ultimately interested in arithmetic Galois groups, the numerical heuristics which we later describe can be a helpful tool.

Finally, the radial camera reconstruction problem of~\Cref{ex:radial} has Galois width $\gr{28},$ compared to a naive degree of $3584.$
This (admittedly extreme) shows the potential payoff of probing the structure of minimal problems through the Galois-theoretic lens.
\end{example}

\section{Numerically Computing Monodromy}

The previous section alluded to an action of a minimal problem's Galois group $\gr{G}$ on the set of solutions $\bl{\pi}^{-1} (\gl{z}) \subset \gr{X} \subset \gl{\CC^m} \times \gr{\CC^n}$ for a generic problem instance $\gl{z}.$
This is the \emph{monodromy action.}
In practice, this may be computed using \emph{parameter homotopy}~\cite[Ch.~7]{sw-book}, a general framework for numerical homotopy continuation methods.
The basic approach requires the following:

\begin{enumerate}
    \item A \emph{fabrication procedure} that produces (numerical approximations of ) generic problem-solution-pairs, $(\gl{z}, \gr{x}) \in \gr{X}$, and
    \item equations $\bl{f_1} (\gr{x} ; \bl{z}), \ldots ,  \bl{f_n} (\gr{x}; \gl{z})$ which locally define $\gr{X}$, ie.
    \[
\operatorname{rank} \left( \displaystyle\frac{\partial \bl{f_i}}{\partial \gr{x_j}}\right) \Bigg\vert_{(\gl{z}, \gr{x})} = \gr{n},
\quad 
\text{ for any generic sample } (\gl{z}, \gr{x}) \in \gr{X}.
    \]
\end{enumerate}

\begin{remark}
The local equations in (2) above may be polynomial or rational functions. See eg.~\cite{DBLP:journals/jsc/DuffR23,simon} for two examples of applications where the latter choice is much more convenient.
\end{remark}

\begin{remark}
If we are given local equations, why not just use Newton's method as the fabrication procedure? 
This often works, but \emph{it is not guaranteed, especially if the vanishing locus $\mathbf{V}(\bl{f_1}, \ldots , \bl{f_n})$ has irreducible components other than $\gr{X}.$ }
When people complain to me that monodromy isn't solving their problem, they are usually either (a) trying to solve a non-minimal problem, or (b) overlooking this particular subtlety.
The takeaway here is: when in doubt, fabricate!
\end{remark}

\begin{remark}
Although minimal problems are locally defined by $\gr{n}$ equations in unknowns, globally one may need more than $\gr{n}$ equations.
This is not really a problem, because the fabrication procedure should ensure that all equations are (approximately) satisfied, and any full-rank square submatrix of the Jacobian yields suitable local equations.
In practice, one should search for such a submatrix carefully---the package \texttt{MonodromySolver}~\cite{duff-monodromy} implements an effective greedy strategy.
\end{remark}

Once the basic requirements (1) and (2) are in place, we are ready to ``do monodromy."
Start by fabricating a problem-solution pair $(\gl{z_0}, \gr{x_0}) \in \gr{X}.$
Then, generate a new generic problem instance $\gl{z_1} \in \gl{\CC^m},$ and two analytic paths $\phi_i : \pur{[0,1]} \to \gl{\CC^m}$, $i=1,2,$ such that the following hypotheses hold:
\begin{enumerate}
\item $\phi_i (\pur{0}) = \gl{z_0}$,  
\item $\phi_i (\pur{1}) = \gl{z_1}$, and
\item for all $\pur{t} \in \pur{[0,1]},$ the fiber $\bl{\pi}^{-1} (\phi_i (\pur{t}))$ consists of $d$ distinct, isolated points.
\end{enumerate}
The path $\phi_0$ determines a \emph{parameter homotopy}
\begin{equation}\label{eq:par-hom}
    \pur{H_0}(\gr{x} ; \pur{t} ) = \begin{pmatrix}
        \bl{f_1} \left( \gr{x} ; \phi_0 (\pur{t})\right) \\
        \vdots \\
           \bl{f_n} \left( \gr{x} ; \phi_0 (\pur{t})\right) 
    \end{pmatrix},
\end{equation}
to which we associate the initial value problem~\cite[Ch.~2]{sw-book}
\begin{equation}\label{eq:davidenko}
\gr{x} ' (\pur{t}) = - \left( \displaystyle\frac{\partial \bl{H_0}}{\partial \gr{x}} \right)^{-1} \, \displaystyle\frac{\partial \bl{H_0}}{\partial \pur{t}}
,
\quad 
\gr{x} (\pur{0}) = \gr{x_0}.
\end{equation}
By our hypotheses on $\phi_0$, this initial value problem admits a local \emph{solution function} $\gr{x} (\pur{t})$, and we may check that $\pur{H} (\gr{x} (\pur{t}); \pur{t})=0$ for all $\pur{t}\in \pur{[0,1]}.$
Thus, numerically integrating~\eqref{eq:davidenko} produces an (approximation of) a solution $\gr{x_1}$ to the problem $\gl{z_1}.$
This is called \emph{path-tracking}, and implemented with a predictor-corrector scheme.

\begin{remark}
By construction, $\gr{x_1}=\gr{x}(\pur{1})$ satisfies the system
\[
\bl{f_1} (\gr{x} ; \gl{z_1}) = 
\ldots 
=
\bl{f_n} (\gr{x} ; \gl{z_1}) =0.
\]
But one can say more: our hypotheses ensure that $(\gl{z_1}, \gr{x_1})\in \gr{X}.$
Nevertheless, this property may silently fail when tracking paths, due to the well-known phenomenon of \emph{path-jumping}.
Thus, it can be helpful to know more about the equations defining $\gr{X}$ globally, or to play with the various path-tracking tolerances.
Another possibility is \emph{certified} path-tracking; based on last year's ISSAC paper~\cite{guillemot}, this should be doable for moderately-sized minimal problems.
\end{remark}

Once we have \emph{two} problem-solution pairs $(\gl{z_0},\gr{x_0}), (\gl{z_1},\gr{x_1})\in \gr{X}$, we can use $\phi_1$ to construct a second parameter homotopy and track the solution $\gr{x_1}$ back to a solution of $\gl{z_0}.$
The pair $(\phi_0, \phi_1)$ determines a \emph{monodromy loop} in the space of problems, that induces a permutation of the solution set of $\bl{\pi}^{-1} (\gl{z_0}).$

Of course, we really want path-tracking to produce \emph{new} solutions, or to somehow be confident that there are none left to find. Here are some practical strategies used for cooking up monodromy loops:
\begin{enumerate}
\item Often (eg.~for all problems in~\Cref{sec:min-probs}), $\gr{X}$ is invariant under a group action $\gr{G} \acts \gr{X}$; fixing $\gr{g_0}, \gr{g_1}\in \gr{G}$, we may take
\begin{equation}\label{eq:segment}
\phi_i (\pur{t}) = (1-\pur{t}) \gl{z_{i}} + \gr{g_i} \cdot \pur{t} \gl{z_{1-i}}.
\end{equation}
Ideally, for generic $\gr{g_0}, \gr{g_1}\in \gr{G}$ our hypotheses on $\phi_i(t)$ are satisfied; a particular case of this is the well-known $\gamma$-trick.~\cite[Ch.~8]{sw-book} Support for constructing this type of monodromy loop is provided by the \texttt{MonodromySolver} option \texttt{Randomizer}.
    \item (``Triangle loops") Rather than chaining two parameter homotopies $\gl{z_0} \leadsto \gl{z_1} \leadsto \gl{z_0}$, one can chain instead chain three parameter homotopies $\gl{z_0} \leadsto \gl{z_1} \leadsto \gl{z_2} \leadsto \gl{z_0}$ using three linear segments like~\eqref{eq:segment}, without the need for a group action.
    \item More generally, the approach of~\cite{duff-monodromy} constructs a \emph{graph of homotopies}, whose cycles can be used to construct generators of the monodromy group. 
    This graph allows loops to be re-used, which is much more efficient than cooking up new loops ``on the fly."
    These ideas have proven useful in both the \texttt{MonodromySolver} implementation and subsequent \texttt{HomotopyContinuation.jl} package~\cite{breiding2018homotopycontinuation}.
\end{enumerate}
\begin{remark}
The ``straight-line" homotopy often used to introduce homotopy continuation is not equivalent to the parameter homotopy~\eqref{eq:par-hom} using the linear segment~\eqref{eq:segment}.
This is because the equations $\bl{f_i} (\gr{x} ; \gl{z})$ may not depend linearly on the parameters $\gl{z}.$
\end{remark}
\begin{remark}
A (non-certified) stopping criterion for tracking monodromy loops can be provided by the trace test.
In its most general form~\cite{trace}, this can be inefficient, as it requires solving another system with many more solutions.
In practice, it is more common to stop once some fixed number of loops fails to make progress.
In the framework of~\cite{duff-monodromy}, one may require that all solutions in the homotopy graph have been tracked along all possible edges.
\end{remark}

The use of the monodromy action in computational algebra has been widespread, with applications like computing Riemann matrices of algebraic curves~\cite{VanHoeij}, polynomial factorization~\cite{GalligoPoteaux}, and numerical irreducible decomposition~\cite{NID}.
An early reference applying these heuristics to compute geometric Galois groups in Schubert calculus is~\cite{anton-schubert}.
Currently, these heuristics are implemented in the publicly-available software packages \texttt{MonodromySolver}, part of the computer algebra system Macaulay2~\cite{M2}, and \texttt{HomotopyContinuation.jl}~\cite{breiding2018homotopycontinuation}.

\begin{remark}
Finally, it is worthwhile to clear up one last occasional misconception: monodromy is mostly useful for constructing \emph{start systems}, not for solving target systems.
The latter are best dealt with via subsequent parameter homotopy runs (eg.~solving a real-valued instance of a minimal problem.)
As~\Cref{ex:radial} illustrates, fully exploiting monodromy loops so as to compute the problem's Galois group can provide a useful tool for optimizing the number of paths tracked by such a homotopy.
\end{remark}

\section{Example with code}\label{sec:example}

Finally, we'll look at the example of the five-point problem discussed in~\Cref{ex:recon}.
We will demonstrate the numerical monodromy heuristic for computing its Galois group, and how to use this information to efficiently solve a real-valued problem instance.
The code has been developed to run in Macaulay2~\cite{M2} (version 1.25.06), and the Galois group is analyzed with GAP~\cite{GAP4} (version 4.14.0.)
I encourage the reader to try running this code, or translating it to work with their preferred software package.

We begin by loading the \texttt{MonodromySolver} package, and setting up variables (solutions), parameters (problems), and equations, namely~\eqref{eq:essential-matrix}.
We assume the fixed camera pair~\eqref{eq:cams-fixed}.

\begin{lstlisting}[language=Macaulay2,basicstyle=\small]
needsPackage "MonodromySolver";
unknowns = gateMatrix{toList vars(t_1,t_2,r_(1,1)..r_(3,3))};
params = gateMatrix{toList vars(p_(1,1,1)..p_(2,5,3))};
R = matrix for i from 1 to 3 list for j from 1 to 3 list r_(i,j);
I = gateMatrix id_(CC^3);
rotationEqs = flatten entries(R * transpose R - I) | {det R - 1};
E = matrix{{0,-1,t_2},{1,0,-t_1},{-t_2,t_1,0}} * R;
pts1 = for i from 1 to 5 list matrix for j from 1 to 3 list {p_(1,i,j)};
pts2 = for i from 1 to 5 list matrix for j from 1 to 3 list {p_(2,i,j)};
ptEqs = apply(pts1, pts2, (p1, p2) -> transpose p2 * E * p1);
\end{lstlisting}

The equations above are implemented as straight-line programs.
This uses the package \texttt{SLPexpressions}, part of a larger constellation of numerical algebraic geometry packages in Macaulay2~\cite{NAG4M2}.
We use the command \texttt{compress} to remove superfluous operations from these straight-line programs, and collect all equations into an instance of the datatype \texttt{GateSystem}.

\begin{lstlisting}[language=Macaulay2,basicstyle=\small]
eqs = gateMatrix{rotationEqs | apply(ptEqs, e -> compress e_(0,0))};
G = gateSystem(params, unknowns, transpose eqs)
\end{lstlisting}
The last line produces the following output:
\begin{lstlisting}[language=Macaulay2,basicstyle=\small]
o12 = gate system: 11 input(s) --> 15 output(s) (with 30 parameters)
o12 : GateSystem
\end{lstlisting}

Next, we implement a procedure that fabricates a generic problem-solution pair.
This relies on Cayley's birational parameterization of $3\times 3$ rotation matrices,
\begin{equation}\label{eq:cayley}
\gr{\CC^3} \ni \gr{\mathbf{t}} \mapsto 
\gr{(I - [\mathbf{t}]_\times ) (I + [\mathbf{t}]_\times ) } \in \gr{\operatorname{SO}_3}.
\end{equation}
\begin{lstlisting}[language=Macaulay2,basicstyle=\small]
fabricate = () -> (
    p1s := apply(5, i -> random(CC^3, CC^1));
    R := random(CC^3, CC^3);
    R = R - transpose R;
    R = (id_(CC^3) - R)^(-1) * (id_(CC^3) + R);
    t := random(CC^2,CC^1) || matrix{{1}};
    p2s := apply(p1s, p -> R * p + t);
    x0 := point{{t_(0,0),t_(1,0)} | flatten entries R};
    z0 := point transpose fold(p1s | p2s, (a, b) -> a||b);
    (z0, x0)
    )
\end{lstlisting}
Next, fabricate a problem-solution pair, and verify that our equations indeed (1) are approximately zero, and (2) locally define  the incidence variety $\gr{X}.$
To aid reproducibility, we set a random seed:
\begin{lstlisting}[language=Macaulay2,basicstyle=\small]
setRandomSeed 2025;
(z0, x0) = fabricate();
norm evaluate(G, z0, x0)
numericalRank evaluateJacobian(G, z0, x0)
\end{lstlisting}

The last two lines generate the following output:
\begin{lstlisting}[language=Macaulay2,basicstyle=\small]
i16 : norm evaluate(G, z0, x0)
o16 = 2.598965467441789e-15
o16 : RR (of precision 53)
i17 : numericalRank evaluateJacobian(G, z0, x0)
o17 = 11
\end{lstlisting}

Since we have more equations than unknowns, we use the function \texttt{squareUp} to obtain a square subsystem of $11$ local equations.
Then, we can finally run monodromy:
\begin{lstlisting}[language=Macaulay2,basicstyle=\small]
Gs = squareUp(z0, x0, G);
(V, npaths) = monodromySolve(Gs, z0, {x0}, NumberOfNodes => 5);
monodromyGroup(V.Graph, FileName => "5pt.g");
\end{lstlisting}

The last line writes a file which can be read by GAP~\cite{GAP4},
containing $19$ permutations in the Galois group of the five-point problem:

\begin{lstlisting}[language=Macaulay2,basicstyle=\footnotesize]
p0:= PermList([18, 5, 15, 10, 3, 11, 8, 9, 19, 6, 20, 13, 14, 1, 17, 7, 4, 16, 12, 2]);
p1:= PermList([5, 3, 6, 1, 2, 16, 11, 4, 17, 9, 18, 14, 12, 8, 13, 15, 19, 20, 7, 10]);
p2:= PermList([1, 6, 11, 2, 5, 3, 16, 14, 10, 4, 7, 8, 13, 9, 15, 12, 18, 17, 19, 20]);
p3:= PermList([18, 11, 8, 16, 2, 13, 10, 5, 1, 7, 15, 9, 12, 6, 17, 4, 3, 14, 19, 20]);
p4:= PermList([16, 7, 17, 19, 1, 13, 14, 5, 9, 3, 11, 10, 15, 18, 4, 20, 6, 8, 12, 2]);
p5:= PermList([1, 11, 3, 2, 16, 6, 10, 8, 5, 7, 13, 9, 4, 14, 15, 12, 18, 17, 19, 20]);
p6:= PermList([3, 6, 5, 20, 16, 18, 10, 17, 1, 7, 15, 8, 4, 13, 14, 19, 2, 12, 9, 11]);
p7:= PermList([6, 9, 10, 2, 19, 1, 5, 15, 16, 13, 4, 11, 20, 7, 8, 12, 18, 17, 14, 3]);
p8:= PermList([5, 3, 6, 1, 2, 16, 11, 4, 17, 9, 18, 14, 12, 8, 13, 15, 19, 20, 7, 10]);
p9:= PermList([9, 16, 18, 7, 15, 13, 8, 5, 3, 6, 14, 4, 1, 17, 11, 10, 19, 20, 12, 2]);
p10:= PermList([15, 9, 14, 18, 8, 7, 16, 10, 12, 4, 2, 11, 6, 3, 1, 17, 13, 5, 20, 19]);
p11:= PermList([15, 13, 2, 11, 8, 18, 3, 17, 4, 14, 16, 5, 6, 12, 1, 9, 19, 20, 10, 7]);
p12:= PermList([15, 14, 12, 4, 19, 18, 13, 17, 7, 5, 10, 3, 20, 2, 1, 16, 8, 6, 9, 11]);
p13:= PermList([6, 11, 2, 5, 1, 14, 10, 3, 4, 7, 16, 9, 15, 12, 8, 13, 18, 17, 19, 20]);
p14:= PermList([4, 20, 17, 2, 5, 7, 11, 10, 1, 9, 15, 19, 13, 18, 16, 12, 3, 14, 6, 8]);
p15:= PermList([1, 17, 9, 7, 2, 8, 3, 6, 19, 14, 20, 18, 12, 11, 15, 10, 16, 4, 5, 13]);
p16:= PermList([20, 7, 8, 5, 2, 1, 3, 15, 4, 14, 16, 10, 12, 6, 19, 13, 9, 11, 17, 18]);
p17:= PermList([15, 14, 6, 19, 18, 13, 2, 5, 4, 12, 16, 3, 17, 8, 1, 20, 11, 9, 10, 7]);
p18:= PermList([2, 16, 20, 9, 18, 3, 6, 14, 7, 8, 10, 4, 17, 19, 12, 11, 5, 13, 15, 1]);
G:=Group(p0, p1, p2, p3, p4, p5, p6, p7, p8, p9, p10, p11, p12, p13, p14, p15, p16, p17, p18);
\end{lstlisting}

We can get some information about this group in a GAP session:

\begin{lstlisting}[language=Macaulay2,basicstyle=\small]
gap> Read("5pt.g");
gap> G;
<permutation group with 19 generators>
gap> StructureDescription(G);
"(C2 x C2 x C2 x C2 x C2 x C2 x C2 x C2 x C2) : S10"
gap> Blocks(G, [1..20]);
[ [ 1, 15 ], [ 2, 12 ], [ 3, 14 ], [ 4, 16 ], [ 5, 13 ], [ 6, 8 ], [ 7, 10 ], [ 9, 11], 
[ 17, 18 ], [ 19, 20 ] ]
\end{lstlisting}

Here is a GAP function that computes the Galois width:

\begin{lstlisting}[language=Macaulay2,basicstyle=\footnotesize]
GaloisWidth := function(G)
  local X, M, C, phi;
  if IsTrivial(G) then return 1;
  elif IsNaturalSymmetricGroup(G) or IsNaturalAlternatingGroup(G) then
    X := OrbitsDomain(G)[1];
    if Length(X) = 4 then return 3;
    else return Length(X);
    fi;
  elif IsCyclic(G) then return Maximum(Factors(Order(G)));
  elif not IsTransitive(G) then return Maximum(List(Orbits(G), O -> 
                              GaloisWidth(Image(ActionHomomorphism(G,O)))));
  else
    X := OrbitsDomain(G)[1];
    if not IsPrimitive(G) then
      phi := ActionHomomorphism(G, Blocks(G, X), OnSets);
      return Maximum(GaloisWidth(Kernel(phi)), GaloisWidth(Image(phi)));
    elif IsSimple(G) then
      M := List(ConjugacyClassesMaximalSubgroups(G), H ->Representative(H));
      return Minimum(List(M, H -> Order(G)/Order(H)));
    else
      C := CompositionSeries(G);
      return Maximum(List([1..Length(C)-1], i -> GaloisWidth(C[i]/C[i+1])));
    fi;
  fi;
end;
\end{lstlisting}

In the \texttt{REPL}, we get the following output:

\begin{lstlisting}[language=Macaulay2]
gap> GaloisWidth(G);
10
\end{lstlisting}

Let me emphasize again that this calculation is largely heuristic.
At best, it tells us only that the group $\gr{G}$ constructed in this example is a subgroup of the true Galois group; moreover, the path-tracking is not certified, which might call into question the veracity of some of the permutations that are constructed.
Nevertheless, experience suggests that the heuristics are pretty reliable.
In the case of the five-point problem, we know that the Galois width must equal $10$, since $20$ solutions divide into $10$ blocks with constant essential matrices.
Going back to Macaulay2, let's see how this information can be used to solve the problem with a $10$-path homotopy.
To compute the action of $\gr{G}$ on the $10$ essential matrices, one may use the decomposable monodromy technique of~\cite{lindberg}.
In \texttt{MonodromySolver}, this is implemented with the option \texttt{Equivalencer} as follows:
\begin{lstlisting}[language=Macaulay2,basicstyle=\small]
W = first monodromySolve(Gs, z0, {x0}, NumberOfNodes => 5, 
            Equivalencer => (x -> point((matrix x)_{0,1})))
\end{lstlisting}
Here, we treat two solutions with the same $\gr{\mathbf{t}}$ as being equivalent: note that specifying $\gr{\mathbf{t}}$ is equivalent to specifying the left-kernel of the essential matrix.

Finally, let us demonstrate how to solve a (randomly-generated) real problem instance with parameter homotopy, using the last monodromy computation to supply the start system:
\begin{lstlisting}[language=Macaulay2,basicstyle=\small]
z0 = matrix W.BasePoint;
x0s = points W.PartialSols;
z1 = random(RR^1, RR^(numParameters Gs));
H = parametricSegmentHomotopy Gs;
H12 = specialize(H, transpose(z0 | z1));
x1s = trackHomotopy(H12, x0s);
\end{lstlisting}
This gives us 10 solutions to the target problem $\gl{z_1}$: the other $10$ solutions may be found by applying the \emph{twisted-pair} symmetry, representing the deck transformation group $\operatorname{Cent} (\gr{G}, S_{20}) \cong \gr{\ZZ_2}$.
(Note: A formula for this symmetry can be recovered using the monodromy-plus-interpolation technique proposed in~\cite{DBLP:conf/issac/DuffKPR23}.

\begin{lstlisting}[language=Macaulay2, basicstyle=\small]
z0 = matrix W.BasePoint;
x0s = points W.PartialSols;
z1 = random(RR^1, RR^(numParameters Gs));
H = parametricSegmentHomotopy Gs;
H12 = specialize(H, transpose(z0 | z1));
x1s = trackHomotopy(H12, x0s);
x1s = x1s | apply(x1s, x -> (
	tvals := (matrix x)_{0,1};
	R := transpose reshape(CC^3, CC^3, (matrix x)_{2..10});
	t := transpose(tvals | matrix{{1}});
	R2 := ((2/(transpose t*t)_(0,0)) * t * transpose t - id_(CC^3)) * R;
	point(tvals | matrix{flatten entries R2})
	)
    );
max apply(x1s, x -> norm evaluate(G, point z1, x))
\end{lstlisting}

The last line outputs a maximum absolute residual on the order of $\approx 10^{-12}$.
Please note that this is the residual for \emph{all $15$ equations}, not merely the $11$ used in homotopy continuation.
Thus, tracking only $\gr{10}$ paths suffices to recover all $20$ solutions with little extra effort. 
In a similar manner, being able to compute the Galois groups, Galois widths, and so on has led to the new, efficient solutions of novel minimal problems surveyed here.

In conclusion, numerical monodromy computation and Galois groups offer a powerful toolkit for solving parametric systems of algebraic equations.
There is more future work that can be done, eg.~concerning tailored data structures~\cite{duff-monodromy,brysiewicz2024monodromy}, parallelization~\cite{DBLP:conf/issac/BlissDLS18,DBLP:journals/ijcse/LeykinV09,DBLP:conf/cvpr/ChienFATTK22}, and certification~\cite{xu2018approach, lee-homotopy,guillemot,van2011reliable}, to name a few directions.

Overall, I feel optimistic that these techniques will continue to be used and improved---both in the setting of minimal problems surveyed here, and more broadly wherever parametric algebraic systems of equations need to be solved.

\bibliographystyle{ACM-Reference-Format}
\bibliography{sample-base}

\end{document}